\pdfoutput=1
\documentclass[11pt]{article}

\usepackage[preprint]{acl}

\usepackage{times}
\usepackage{latexsym}
\usepackage{amsmath}
\usepackage{graphicx}
\usepackage{listings}

\usepackage{algorithm2e}
\usepackage{tcolorbox}
\usepackage{lscape}
\usepackage[table]{xcolor}
\usepackage[T1]{fontenc}
\usepackage[utf8]{inputenc}
\usepackage{microtype}

\usepackage{inconsolata}

\usepackage{amssymb}
\usepackage{subcaption}
\usepackage{multirow}
\usepackage{changepage}
\usepackage{hhline}
\usepackage{color}
\usepackage{booktabs}
\usepackage{bbm}
\usepackage{wrapfig,lipsum,booktabs}
\usepackage{makecell}
\usepackage{float}
\usepackage{arydshln}
\usepackage{enumitem}
\usepackage{dsfont}
\usepackage{bbding}
\usepackage{authblk}

\title{How do Language Models Generate Slang: A Systematic Comparison between Human and Machine-Generated Slang Usages}

\author{
 \textbf{Siyang Wu\textsuperscript{1, 2}},
 \textbf{Zhewei Sun\textsuperscript{2}}
\\
\\
 \textsuperscript{1}Data Science Institute, University of Chicago, Chicago, Illinois\\
 \textsuperscript{2}Toyota Technological Institute at Chicago, Chicago, Illinois\\
 \ttfamily siyangwu@uchicago.edu, zsun@ttic.edu
}

\begin{document}
\maketitle
\begin{abstract}

Slang is a commonly used type of informal language that poses a daunting challenge to NLP systems.
Recent advances in large language models (LLMs), however, have made the problem more approachable. 
While LLM agents are becoming more widely applied to intermediary tasks such as slang detection and slang interpretation, their generalizability and reliability are heavily dependent on whether these models have captured structural knowledge about slang that align well with human attested slang usages.
To answer this question, we contribute a systematic comparison between human and machine-generated slang usages. Our evaluative framework focuses on three core aspects: 1) \textit{Characteristics} of the usages that reflect systematic biases in how machines perceive slang, 2) \textit{Creativity} reflected by both lexical coinages and word reuses employed by the slang usages, and 3) \textit{Informativeness} of the slang usages when used as gold-standard examples for model distillation.
By comparing human-attested slang usages from the Online Slang Dictionary (OSD) and slang generated by GPT-4o and Llama-3, we find significant biases in how LLMs perceive slang.
Our results suggest that while LLMs have captured significant knowledge about the creative aspects of slang, such knowledge does not align with humans sufficiently to enable LLMs for extrapolative tasks such as linguistic analyses.
%, substantial differences in how creative they are, and that machine-generated slang tend to be less informative when used for fine-tuning.
%Our results suggest that while current LLMs can be effective in simple tasks, they have not yet fully captured structural knowledge about slang to enable their use in complex generative tasks and linguistic analyses.

%Specifically, the way in which machines define and use slang can differ from humans in multiple dimensions.

% it is unclear how such models perceive slang and how well do they align with humans. 

% While LLMs become increasingly more capable in processing slang

% Before we can apply these models in the wild, it is important to understand the 

% While recent work has shown that LLMs trained on large text corpora brings a generational leap in downstream performance on slang-related tasks, it is unclear what kinds of structural knowledge about slang these models capture.

\end{abstract}

\section{Introduction}

Slang is a type of informal language that is commonly used in colloquial speech~\cite{sornig81}. The use of slang is both creative~\cite{warren92, eble12} and ephemeral~\cite{eble89}, meaning that slang is not only more difficult to comprehend compared to conventional language, but it is also necessary to handle an ever-evolving repertoire of novel slang usages. Such characteristics of slang necessitate natural language processing (NLP) systems that can adapt to unseen slang usages without expansive retraining. While earlier work on NLP for slang often resorts to retrieval-based systems that can hardly generalize (e.g., \citealp{pal13}), recent work has been successful in developing systems for the automatic detection~\cite{pei19, liu21, sun24}, generation~\cite{kulkarni18, sun21}, and interpretation~\cite{ni17, sun22, mei24, wuraola24} of slang that can generalize well toward novel slang usages that have never been seen during training. In particular, large language models (LLMs) have been very effective in many tasks involving slang under both zero-shot and few-shot settings, suggesting that the LLMs have, to some extent, captured structural knowledge about slang that enables generalization.

\begin{figure}[t]
  \centering
  \includegraphics[width=\columnwidth]{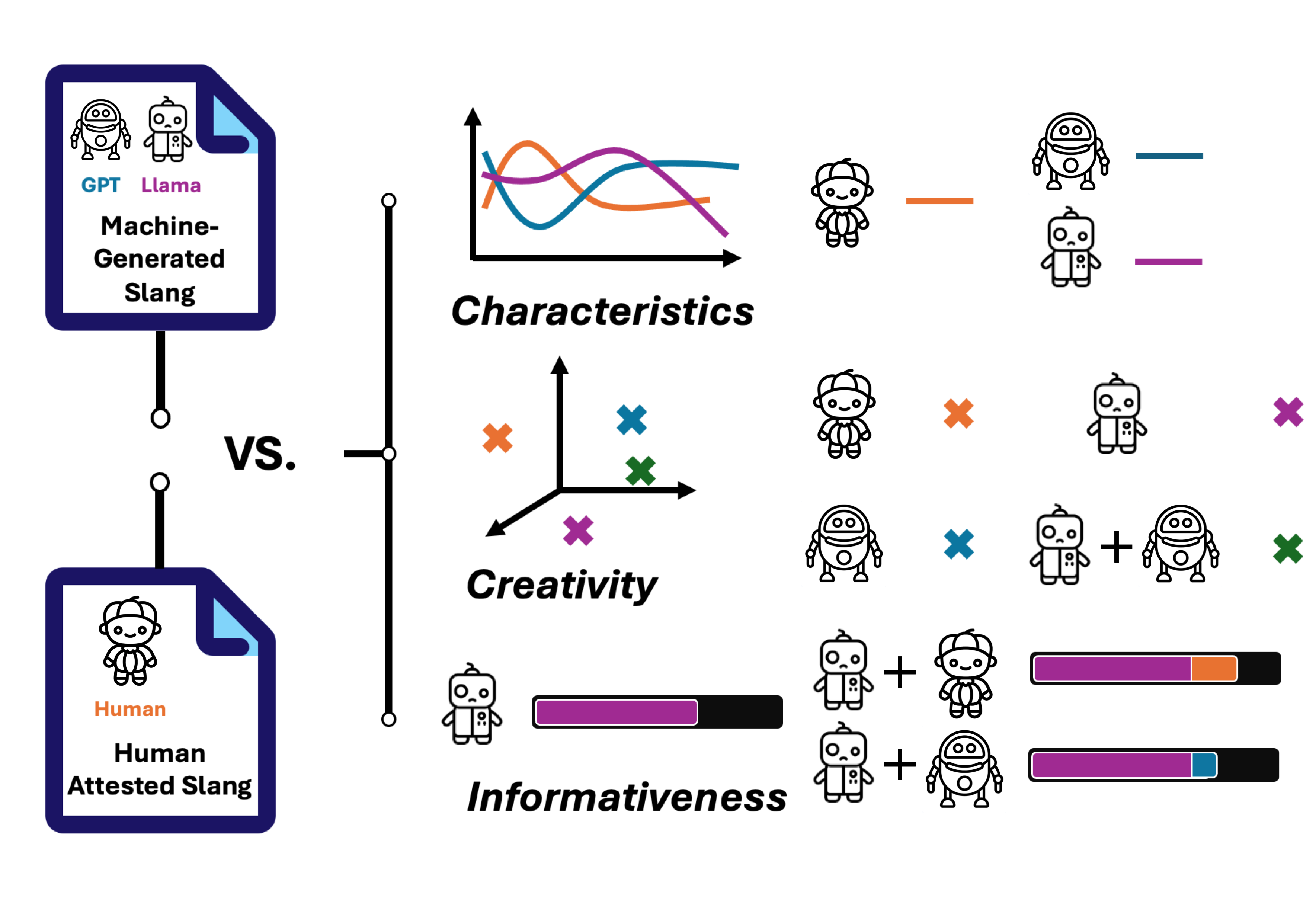}
  \caption{Our evaluative framework considers three core aspects of knowledge: 1) Characteristics, 2) Creativity, and 3) Informativeness by comparing human and machine-generated slang usages. The + sign indicates fine-tuning.}
  \label{fig1}
\end{figure}

It is not well understood, however, what underlying structures about slang the LLMs have captured and whether they are comparable to human knowledge. Such insights are valuable in gauging the reliability of downstream applications that require a precise characterization of slang. For example, a model with misaligned knowledge would be systematically biased toward detecting certain types of slang usages. Slang usages generated by such models can also misinform downstream agents that consume the content (e.g., model distillation), as well as linguistic analyses that rely on LLMs for automatic annotation.

We make a first step toward the interpretation of LLMs' internal knowledge about slang by collecting a dataset of machine-generated slang usages from GPT-4o~\cite{openai24} and Llama-3~\cite{llama24} that span a diverse set of controlled conditions. % elaborate here
Specifically, we prompt the LLMs to generate novel slang usages that each encompasses 1) A slang term, 2) A sense definition sentence, and 3) A usage context. We perform both controlled generation where the model is provided with existing senses from a slang dictionary and an uncontrolled setting where the model is allowed to generate slang usages attached to any senses. Under each condition, we further constrain the type of word choices that can be made. This includes 1) \textbf{Lexical coinage} where the the model is prompted to create a new term, 2) \textbf{Word reuse} where the model chooses an existing term in the lexicon, and 3) \textbf{Free-form} generation without any restrictions. 
We collect at least 1,000 valid generations in each setting for a total of 58,197 machine-generated slang usages\footnote{Code and data available at: \url{https://github.com/siyangwu1/LLM-Slang-Dictionary}}.

Using this dataset, we make a systematic comparison between human-generated slang usages attested by slang dictionaries and machine-generated slang usages. 
Illustrated in Figure~\ref{fig1}, we propose an evaluative framework that makes comparisons along three core aspects: 1) Our framework compares the aggregate usage  \textbf{characteristics} of the generated slang usages to discern any systematic biases in how machines perceive slang; 2) It measures and compares the \textbf{creativity} of slang usages, examining morphological complexity and semantic coherence of coined terms as well as semantic novelty and contextual surprisal for cases of reuse; 3) Model distillation is performed using both human and machine-generated slang usages as examples to measure their \textbf{informativeness} toward a diverse set of NLP tasks for slang.
Our results show that while LLMs have captured sufficient creative knowledge to generate plausible slang usages, the generated usages still deviate significantly from human-generated usages in certain aspects.
%Our results show that while LLMs can generate plausible slang usages, the generated usages still deviate significantly from human-generated usages across all three aspects.

We make the following contributions in this paper: 1) A dataset of slang usages generated by GPT-4o and Llama-3 that enables studies of machine-generated informal language; 2) An evaluative framework that assesses knowledge alignment between human and machine-generated language use; 3) The first systematic comparison between human and machine-generated slang usages.

\section{Related Work}

\subsection{Knowledge-driven processing of slang}

Earlier work relies on building and retrieving from high quality data sources to enable machine processing of slang~\cite{pal13, dhuliawala16, gupta19}. Such approaches require constant updates to obtain new knowledge and thus cannot be efficiently combined with large machine learning models that are expansive to retrain. Meanwhile, encoder models such as BERT~\cite{devlin19} perform poorly on slang due to limited scale at the time. To address this, a series of work has been proposed to inject linguistic knowledge about slang into the models to create inductive biases that enable efficient learning. Such an approach has been successfully applied to enable automatic detection~\cite{pei19, liu21}, generation~\cite{kulkarni18, sun19, sun21}, and interpretation~\cite{sun22} of slang usages that have not been seen during training.

Most related to our work is \citet{kulkarni18} who built generative neural models of slang word coinage for common types of word formation strategies (e.g., blending) employed in slang word formation identified by prior linguistic research~\cite{mattiello13}. Also, \citet{sun21} studied cases of word reuse in slang by modeling slang generation as a sense extension phenomenon. In their work, contrastive learning~\cite{baldi93, bromley94, weinberger09} was applied to construct a sense embedding space that encapsulates commonly used sense extension patterns (e.g., bad to good) in slang. In our work, we are interested in identifying whether the machine-generated slang usages adhere to such linguistic structures that are informative when modeling both cases of lexical coinage and word reuse.

\subsection{Language modeling and slang}

As an alternative to knowledge-driven approaches, several studies have explored the possibility of learning knowledge about slang directly from large scale text corpora. Earlier work adopted sequence models for both automatic slang interpretation~\cite{ni17} and generative word formation~\cite{wibowo21}. In both cases, the models require a large set of task-specific training data to achieve adequate performance. Recent advances in LLMs have alleviated the need to provide task-specific training data. \citet{sun24} evaluated GPT-4~\cite{openai23} on both slang detection and the inference of a slang's demographic origin in zero-shot settings. While task-specific training datasets were still shown to be useful, the zero-shot models show comparable performance with BERT-like models that have been fine-tuned on task-specific data. \citet{wuraola24} applied ChatGPT-4~\cite{openai23}, Gemini~\cite{gemini24}, and Llama-3-8B~\cite{llama24} to slang interpretation and also achieved good performance. \citet{mei24} showed that causal inference techniques can be wrapped around LLMs to make further improvements to their predictive accuracies on slang interpretation when compared to traditional prompting methods (e.g., \citealp{wei22}).

Recent progress in this field suggests that LLMs have captured structural knowledge about slang to some extent that enables them to process slang effectively. We extend this line of work by critically examining the prospect of applying LLMs to more complex generative tasks and linguistic analyses involving slang.

\section{Data}

We first collect sets of slang usages generated by both humans and machines. We use attested slang usages from the Online Slang Dictionary (OSD)\footnote{\url{http://onlineslangdictionary.com/}} as the set of human-generated slang. In OSD, each slang usage consists of:

\begin{enumerate}
    \item \textbf{Lexical term}: A word or phase that denotes slang usage. E.g., "bruddah".
    \item \textbf{Definition sense}: A definition sentence for the slang sense attached to the lexical term. E.g., "Alternate spelling of brother".
    \item \textbf{Usage context}: A sentence capturing the context in which the slang is being used. E.g., "Safe, my \textit{bruddah}".
\end{enumerate}

We obtain 9,115 slang usage entries from \citet{sun21}'s OSD dataset by sampling one usage from each unique term. We use OSD entries as our human-generated slang baseline when comparing with machine-generated slang entries in our experiments. In our results, we also use the English Wiktionary~\cite{ylonen22} to partition OSD entries into three categories based on degree of conventionalization: 1) Highly conventionalized slang (High-Conv) with a matching lexical entry in Wiktionary containing a definition that is not tagged as informal\footnote{We consider a Wiktionary definition to be informal if it is attached with one of the following tags: \{slang, informal, jargon, idiomatic\}.}; 2) Conventionalized slang (Conv) where the slang has a matching informal definition; and 3) Non-conventionalized slang (Non-Conv) where no matching definition exists. We consider two definition sentences to be close matches if at least 50\% of their content words overlap.

For machine-generated slang, we use GPT-4o and Llama-3-8B to each generate a set of slang usages. We consider two generation settings. First, we perform \textbf{controlled} generation where each generation prompt is conditioned on an existing definition from OSD. Here, The model is asked to assign a slang term that would express the given human-defined meaning along with an example usage context. This setup focuses on making word choices grounded in existing concepts. We also perform \textbf{uncontrolled} generation where the model is able to express concepts outside ones attested in the slang dictionary. In this case, the generated usages rely solely on the model's intrinsic knowledge about slang obtained during pre-training. Under each setting, we further control for the type of word choice made by the model. Under the \textit{Coinage} condition, the model is prompted to generate a novel term. Conversely, under the \textit{Reuse} condition, the model is prompted to assign an existing term. Finally, we include a \textit{Free-form} condition where the model can pick either types. We ensure compliance by checking the model generated terms against the English Wiktionary~\cite{ylonen22}\footnote{See details in Appendix~\ref{sec:slangGenerationPsudocode}.}. We filter out all model outputs that do not conform to the instructions. Table~\ref{tab:generation_schema} summarizes all control conditions used for generation and the corresponding data partitions. 
We will refer to the partitions outlined in Table~\ref{tab:generation_schema} throughout the paper.

Under the uncontrolled condition, we generate slang usages iteratively until we reach 1,000 usage entries. % (see Appendix~\ref{sec:slangGenerationPsudocode} for our generation algorithms). 
To enhance the diversity of the generated slang usages, we first implemented the method proposed by~\citet{chen24} which introduces prompt-level randomness by prompting with a randomly generated number attached to each instance of generation. However, it yielded suboptimal results compared to the baseline (see Appendix~\ref{app-gendiver} for an ablation). As a result, we adopted a simple sampling configuration with a temperature of 1.2 and top-p of 0.95. We use default values for all other hyper-parameters. This best-performing configuration yielded 765 unique compliant entries out of the first 1,000 generations.

To remove duplicates in generation, we ensure that each generated item must contain either a unique slang term or sense definition. We allow duplicate terms only if their associated senses are semantically distinct, defined as having a cosine similarity below 0.8 between their corresponding \textit{all-MiniLM-L6-v2} Sentence-BERT (SBERT, \citealp{reimers19}) embeddings.

To avoid generating duplicates over overlapping senses in the controlled generation setting, we apply clustering to all definition sense embeddings. Specifically, we encode all sense definition sentences from OSD using SBERT and apply clustering with DBSCAN \cite{DBSCAN} using default hyperparameters. Out of the 9,115 usage entries from OSD, the DBSCAN clustering procedure yielded 7,890 distinct word sense clusters.

For each sense cluster $\mathcal{C} = \{s_1, s_2, \dots, s_n\}$, we choose the most frequent sense definition sentence $s^*$ to be used in the prompt. If necessary, we break ties by random sampling.
Given $s^*$, we prompt the LLM to generate a list of terms $\mathcal{T} = \{t_1, t_2, \dots, t_n\}$ and contexts $\mathcal{C} = \{c_1, c_2, \dots, c_n\}$ such that each term $t_i$ and context $c_i$ defines a slang usage with definition $s^*$. We perform the generation iteratively and reject any duplicated terms that have already been generated until we have n generated usages.
% Specifically, we enforce the following constraint:
% \begin{align}
% \cos(f(w_i), f(w_j)) < 0.8 \quad \text{for } i \neq j.
% \end{align}
% Here, $f(\cdot)$ denotes the XXX embedding of the word. 
The prompts used for data collection are included in Appendix~\ref{app:slang-prompt}. The algorithms used for the prompting and filtering processes are in Appendix~\ref{sec:slangGenerationPsudocode}. Examples from both OSD and our generated datasets can be found in Appendix~\ref{app-sampledata}.

%\subsection{Human-generated slang} \label{data-human}

%\subsection{Machine-generated slang} \label{data-machine}

% We design our generation schema following the structure summarized in Table~\ref{tab:generation_schema}, which covers three generation types (\textit{free}, \textit{reuse}, and \textit{coinage}) under two settings: \textit{uncontrolled} and \textit{controlled}.

\begin{table}[t]
  \centering
  % \resizebox{\columnwidth}{!}{%
    \begin{tabular}{lccc}
      \textbf{Setting} & \textbf{Free-form} & \textbf{Reuse} & \textbf{Coinage} \\
      \hline
      \addlinespace[0.05cm]
      Uncontrolled      & U-F & U-R & U-C \\
      Controlled  & C-F & C-R & C-C \\
    \end{tabular}
  % }
  \caption{Data partitions for slang usage generation under all possible experimental conditions.}
  \label{tab:generation_schema}
\end{table}

\section{Experiments}
\subsection{Characteristics} \label{sec_characteristics}

\begin{figure}[t]
  \centering
  \includegraphics[width=\columnwidth]{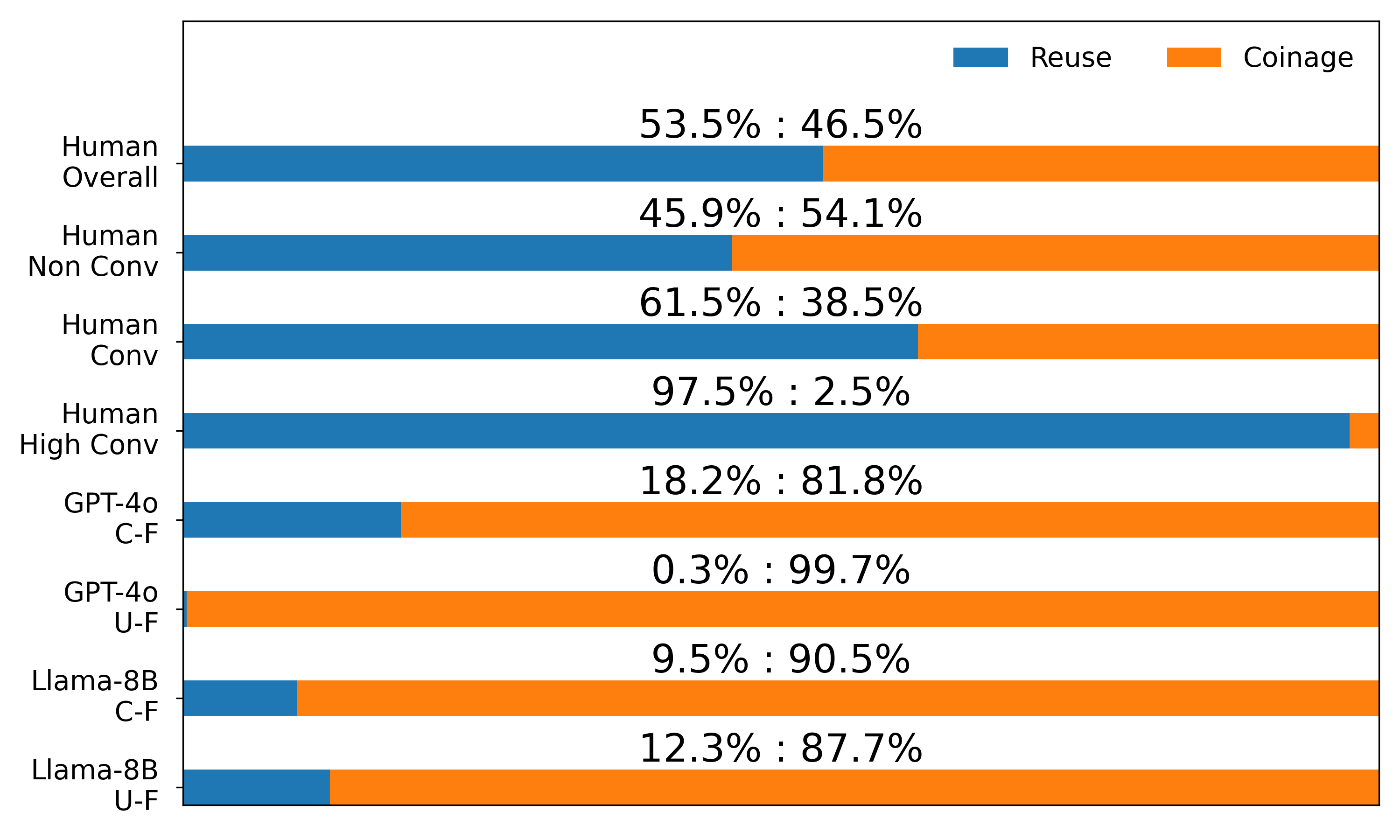}
  \caption{Reuse–Coinage proportion of human (OSD) vs machine-generated slang.}%{\color{blue} The Instruct model is used for LLaMA-8B.}}
  \label{fig:reuse_prop}
\end{figure}

We first compare and examine the aggregate characteristics of human and machine-generated slang usages with respect to usage types, word formation patterns, and expressed topics. Figure~\ref{fig:reuse_prop} shows the distribution of the human and machine-generated slang across two usage types: lexical coinage and word reuse. We observe clear distinctions between slang usages from OSD and the LLMs. While OSD exhibits a more balanced distribution across the two usage types, both GPT-4o and Llama-3 display a strong inclination toward producing coinages. 
In the uncontrolled case, GPT-4o only produced 3 coinages out of 1,000 generations.
Interestingly, prompting GPT-4o using slang senses from human attested usages (i.e., controlled case) significantly reduces the imbalance. %by generating many more cases of word reuses. 
The proportion of reuse cases, however, is still much lower compared to human-generated OSD. The machine-generated slang shows that LLMs' perception of slang is heavily biased toward the use of novel terms.

\begin{figure}[t]
  \includegraphics[width=\columnwidth]{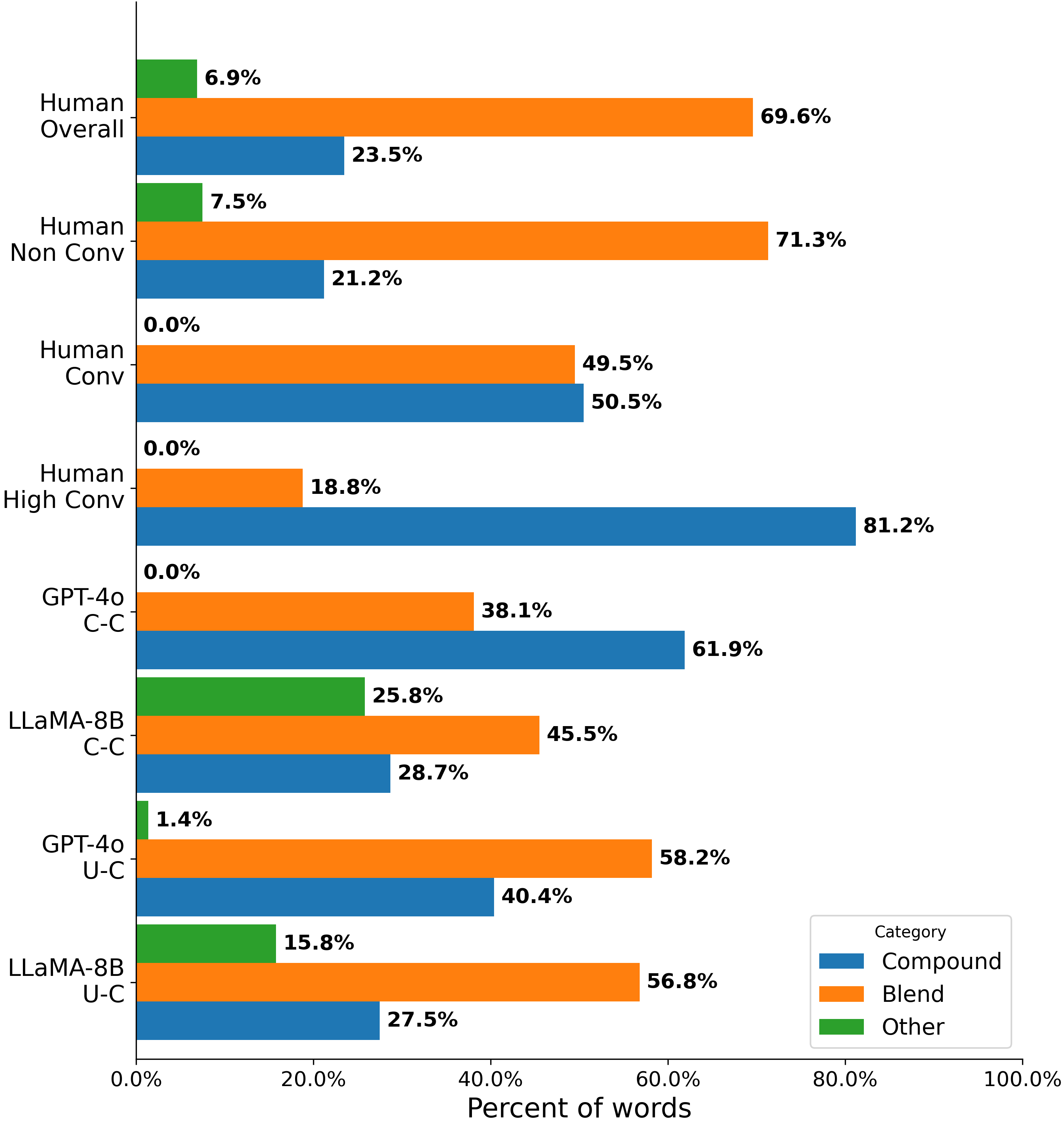}
  \caption{Distribution of word formation processes used in both human and machine-generated slang.}
  \label{fig:coinage_class}
\end{figure}

\begin{table}[t]
\centering
\scalebox{0.7}{%
\begin{tabular}{lrr}
%\hline
\textbf{Topic} & \textbf{OSD Topic Words} & \textbf{GPT-4o Topic Words} \\
\hline
\addlinespace[0.05cm]
Topic 1 & find, people, area & movement, energetic, digital \\
%\hline
Topic 2 & acronym, sexual, p**is & energy, excitement, enthusiasm \\
%\hline
Topic 3 & person, money, f*** & playful, laughter, fleeting \\
%\hline
Topic 4 & sex, sh**, spell & unexpected, surprise, reaction \\
%\hline
Topic 5 & female, term, attractive & quick, smile, attention \\
%\hline
\end{tabular}
}
\caption{Representative topic words from both OSD and GPT-4o sense definitions. Full results are in Appendix~\ref{sec:Topic analysis Full table}.}
\label{tab:topic_words}
\end{table}

For cases of coinage, we also examine the word formation processes that are employed. In the case of human-generated usages, a set of common word formation processes have been identified in the literature~\cite{mattiello13}. One of the most prominent processes is blending, where parts of two lexical items are combined to create a new term (e.g., \textit{lambortini}). Related to blending is the process of compounding~\cite{lehrer70, traugott05} in which two words are combined verbatim (e.g., \textit{backwash}).
Here, we approximate the proportion of the coinage cases that employs either compounding or blending using lexical decomposition (see detailed algorithm in Appendix~\ref{sec:CoinageCategoryClassificationPsudocode}). First, we use Morfessor~\cite{smit-etal-2014-morfessor} to perform morphological segmentation on each slang term and only consider words with more than two morphological segments in this experiment. We then query each of the two segments in Wiktionary.
% For all words that decompose to exactly two segments\footnote{We only consider words with exactly two morphological segments in this experiment.}, we query these segments in Wiktionary. 
If exact matches can be found for all segments, we label the term as a compound. Otherwise, if a segment can be found as a prefix of a word in Wiktionary and the next segment can be found as a suffix then the word is labeled as a blend. All other types of coinages are labeled as 'Other'\footnote{See Appendix~\ref{Additional Examples} for examples that fall under each word formation category.}.
We apply the labeling routine to all coinage cases in OSD and both the controlled and uncontrolled machine-generated slang usages generated under the coinage condition. Figure~\ref{fig:coinage_class} shows the proportion of coinage cases that have been labeled into each word formation categories. We observe similar distributions from the uncontrolled GPT-4o generations and human data with GPT-4o biased toward more compounds than blends. Interestingly, when we control GPT-4o using definitions from OSD, the model shows an even stronger bias toward coining compound words. Llama-generated usages, on the other hand, show much less preference toward either compounding or blending compared to both human and GPT-generated slang. The distinction in these distributions shows that individual LLMs obtain its own perception of slang that is not necessarily well-aligned with human knowledge.
%For both sets of GPT-4o generated usages, we observe few to none cases assigned to the 'Other' category which contains strategies such as reduplication and acronyms. 
%Overall, we observe notable distinction between the word formation strategies employed by humans compared to LLM-generated slang.

Finally, we examine topical preferences in the slang usages by applying LDA topic modeling~\cite{blei03} on 1,000 definition sentences (filtering out all stop words) each from both OSD and the U-F set from both LLMs. We use Gensim~\cite{rehurek_lrec} to extract the 20 most representative words under 5 topics. 
%We apply topic modeling .
Table~\ref{tab:topic_words} shows example words from each topic. Consistent with previous findings~\cite{labov72, labov06}, our results show a strong preference toward taboo topics such as sex and profanity in real slang from humans. Meanwhile, machine-generated slang shows a notable preference toward more positive but less concrete concepts. One hypothesis is the influence of alignment techniques (e.g., RLHF; \citealp{ouyang22}) prevents the models from producing outputs involving potentially offensive or controversial content and instead steers them toward neutral or positive expressions. Our results reaffirm that human slang are created to reflect cultural dynamics while suggesting that LLMs' generations merely capture the creative aspect of the generative process.

\begin{table}[t]
\centering
\small
% \resizebox{\columnwidth}{!}{%
\begin{tabular}{lrr}
%\hline
\textbf{Source} & \textbf{Mean} & \textbf{Std} \\
\hline
\addlinespace[0.1cm]
Human (OSD)                                & 2.032 & 0.841 \\
~ - Non-Conv                                & 2.051 & 0.909 \\
~ - Conv                                & 1.941 & 0.721 \\
~ - High-Conv                                & 1.909 & 0.944 \\
\addlinespace[0.05cm]
GPT-4o C-C                          & 2.442 & 0.797 \\
GPT-4o U-C                          & 2.634 & 0.655 \\
\addlinespace[0.05cm]
Llama-3-8B-INT                       & 1.698 & 0.741 \\
Llama-8B + GPT4o C-C        & 1.985 & 0.807 \\
Llama-8B + GPT4o U-C        & 2.038 & 0.748 \\
Llama-8B + OSD-C             & 1.811 & 0.794 \\

%\hline
\end{tabular}
% }
\caption{Morphological complexity scores for coined terms measured by the number of segments in their respective Morfessor decompositions.}
\label{tab:source_coinage_mean_std}
\end{table}

\begin{table}[t]
\centering
\resizebox{\columnwidth}{!}{%
\begin{tabular}{lrrrr}

\textbf{Source} & \textbf{Mean} & \textbf{Std} & \textbf{IQR} & \textbf{Kurtosis} \\
\hline
\addlinespace[0.1cm]
Human (OSD)                      & 1.286 & 0.058 & 0.076 & 0.185 \\
~ - Non-Conv                        & 1.289  & 0.060  & 0.072 & 3.852 \\
~ - Conv                        & 1.294  & 0.055  & 0.065    & 0.731 \\
~ - High-Conv                        & 1.280  & 0.076 & 0.063 & -0.580 \\
\addlinespace[0.05cm]
GPT‑4o C-C                     & 1.277 & 0.055 & 0.069 & 0.275 \\
GPT-4o U-C                     & 1.250 & 0.058 & 0.076 & 0.136 \\
\addlinespace[0.05cm]
Llama-3-8B-INT                   & 1.290 & 0.061 & 0.085 & 1.180 \\
Llama-8B + GPT-4o C-C     & 1.291 & 0.059 & 0.075 & 0.270 \\
Llama-8B + GPT-4o U-C    & 1.274 & 0.055 & 0.074 & 0.219 \\
Llama-8B + OSD-C           & 1.308 & 0.049 & 0.065 & 1.957 \\

\end{tabular}
}
\caption{Morphological coherence scores for compound words across all sources. Lower values indicate better semantic alignment between the slang senses and the coined terms.}
\label{tab:coherence_scores}
\end{table}

\subsection{Creativity} \label{sec-creativity}

\subsubsection{Creativity in Coinage}
%----------------------------------------------------------new-----------------------------------------------
We evaluate the morphological creativity of coined slang terms through two key aspects: \textit{morphological complexity} and \textit{morphological coherence}. We define morphological complexity as the average number of morphological segments a coined term has. Here, higher complexity reflects more elaborate word-composition strategies being employed. Our measure of word complexity is largely motivated by psycholinguistic work studying how lexical complexity affects efficient communication (e.g.,~\citealp{doi:10.1073/pnas.2025993118,doi:10.1073/pnas.2406971121}). Existing findings suggest that word complexity is a contributing factor in enabling efficient communication and that competing pressures need to exist for humans to prefer more complex word forms. In our context, we attribute the need to create more complex words as a mechanism for illustrating creativeness in novel slang usages. 

We also measure the morphological coherence of each coined compound by comparing semantic representations of the slang sense with senses corresponding to each constituent word. Our coherence metric is analogous to topic coherence measures that assess semantic consistency using word embeddings~\cite{10.1145/2911451.2914729}. Concretely, we define coherence for newly coined compounds as the average similarity between the embedding of the compound’s definition and the embeddings of its constituent subword definitions. Here, better coherence suggests that the coined term is more semantically grounded with respect to its morphological structure. While morphological segmentation reflects surface-level complexity, coherence measures whether the meaning of the coined word is semantically consistent with its constituent morphological segments, thus reflecting a more nuanced level of creativity.

%-------------------------------------------------------------------old ---------------------------------
% We evaluate the morphological creativity of coined slang terms through two key aspects: \textit{morphological complexity} and \textit{morphological coherence}. We define morphological complexity as the average number of morphological segments a coined term has. Here, higher complexity reflect more elaborate word composition strategies being employed. 
% We also measure the morphological coherence of each coined compound by comparing semantic representations of the slang sense with senses corresponding to each constituent word.
% Here, better coherence suggests that the coined term is more semantically grounded with respect to its morphological structure.
% While morphological segmentation reflects surface-level complexity, coherence measures whether the meaning of the coined word is semantically consistent with its constituent morphological segments, thus reflecting a more nuanced level of creativity.

\paragraph{Morphological Complexity.}

We use Morfessor to decompose all slang terms that are not found in Wiktionary. Table~\ref{tab:source_coinage_mean_std} presents the average number of morphological segments per coined term in both OSD and machine-generated coinages. We find that GPT-coined terms are much more complex compared to coinages from OSD, for both the controlled and uncontrolled conditions, especially in the uncontrolled case where the model also shows significantly lower variability in complexity. Meanwhile, The Llama model produces much simpler constructions compared to both OSD and GPT-4o, suggesting that different LLMs have varied preferences toward lexical creativity in slang.

\paragraph{Morphological Coherence.}

We compute the morphological coherence score for all compounds by taking the average Euclidean distance between the SBERT embedding of the coined term's slang definition and the embeddings of sense definitions corresponding to the term's constituent words.
Table~\ref{tab:coherence_scores} reports the results. 

Among the models, GPT-4o under the uncontrolled setting achieved the lowest mean score, indicating that its coined terms are not only morphologically complex but also more semantically coherent. In contrast, coherence degrades notably when GPT-4o is conditioned on human attested senses (C-C), reflecting a reduction in semantic consistency. The results indicate that GPT-4o prefers more semantically coherent word choices when coining new words while humans are more playful in their word choices.

\subsubsection{Creativity in Reuse}
%------------------------------------------------new------------------------------------------------------
Aside from lexical coinage, slang also employs flexible reuse of existing words in creative ways~\cite{warren92}. We measure creativity in two distinct aspects. First we measure the \textit{novelty} of the slang sense extension encompassed by the word choice in a reuse. Motivated by computational models of slang reuse~\cite{sun21}, we measure novelty by computing the semantic distance between the intended sense $S$ of the slang term with the prototypical sense representation~\cite{rosch75} of the reused term $t$:
\begin{align*}
    \text{Novelty}(t, S) = \| E(S) - \frac{1}{|\mathcal{C}_t|}\sum_{i=1}^{|\mathcal{C}_t|}E(\mathcal{C}_{t_i}) \|_2
\end{align*}
Here, $\mathcal{C}_t$ denotes the set of existing senses of the term $t$ in a conventional dictionary and $E(\cdot)$ is an embedding function. The prototypical sense representation encapsulates the aggregate meaning of $t$ and thus the difference between it and the slang sense representation reflects novelty of the sense extension. 

This approach follows well-established measures in cognitive science that model word sense extension and linguistic categorization~\cite{doi:10.1073/pnas.1714730115,Habibi2020-HABCAT-4}. Under these frameworks, a shorter semantic distance implies less creativity and would thus make an extension easier for the human brain to process. Such metrics have also been applied to model creative language use such as metaphor~\cite{stowe-etal-2021-metaphor} and slang~\cite{sun21}.

We compute an alternative measure of creativity in reuse inspired by the Cooperative Principle of \citet{grice1975logic} and the Principle of Relevance proposed by \citet{sperber1986relevance}. Under these frameworks, the lack of relevance in the use of a lexical item indicates the speaker's intention to invoke a novel or enriched meaning. In extension, we infer the creativity of a slang reuse by computing its \textit{surprisal} in context. Surprisal is a well-established measure that correlates with behavioral and neural indicators of human processing effort~\cite{DEMBERG2008193,SMITH2013302,goodkind-bicknell-2018-predictive,wilcox2020predictivepowerneurallanguage,LI2023105359}. Recent findings further validate that surprisal computed from LLMs aligns with human data and significantly improves the prediction of reading time over baseline models~\cite{nair-resnik-2023-words}. Prior work in creative NLP has also applied surprisal as a principle behind pun generation~\cite{he-etal-2019-pun}. More broadly, the use of surprisal in modeling creativity is motivated by incongruity theory~\cite{Veale+2004+419+428}, which posits that humor and creativity are perceived when there are mismatches between two instantiations of words (in our case, between the lexical meaning and the contextual meaning of a word

We operationalize surprisal by computing the average negative log-likelihood score assigned to all constituent tokens of a slang term by the LLM conditioned on the preceding context. Higher surprisal indicates that the model finds the term less predictable in the given context. The surprisal score thus measures the creativity of the slang usage with respect to the usage context.

\begin{table}[t]
\centering
\resizebox{\columnwidth}{!}{%
\begin{tabular}{lrrrr}
%\hline
\textbf{Model} & \textbf{Mean} & \textbf{Std} & \textbf{IQR} & \textbf{Kurtosis} \\
\hline
\addlinespace[0.1cm]
Human (OSD)                       & 1.141 & 0.209 & 0.220 & 4.591 \\
~ - Non-Conv                       & 1.180 & 0.162 & 0.184 &  2.037 \\
~ - Conv                       & 1.098 & 0.250 & 0.295 &  2.675 \\
~ - High-Conv                       & 0.966 & 0.294 & 0.407 &  1.156 \\
\addlinespace[0.05cm]
GPT-4o C-R                        & 1.231 & 0.127 & 0.130 & 6.870 \\
GPT-4o U-R                        & 1.226 & 0.107 & 0.118 & 4.658 \\
\addlinespace[0.05cm]
Llama-3-8B-INT                    & 1.222 & 0.124 & 0.144 & 3.190 \\
Llama-8B + GPT4o C-R      & 1.257 & 0.108 & 0.123 & 4.095 \\
Llama-8B + GPT4o U-R      & 1.252 & 0.104 & 0.121 & 2.790 \\
Llama-8B + OSD-R            & 1.257 & 0.106 & 0.127 & 5.691 \\

%\hline
\end{tabular}
}
\caption{Summary of novelty statistics across all cases of word reuse.}
\label{tab:novelty_stats}
\end{table}
\begin{table}[t]
\centering
\resizebox{\columnwidth}{!}{%
\begin{tabular}{lrrrr}
%\hline
\textbf{Source} & \textbf{Mean} & \textbf{Std} & \textbf{IQR} & \textbf{Kurtosis} \\
\hline
\addlinespace[0.1cm]
Human (OSD)                    & 28.73 & 13.06 & 17.59 &  0.73 \\
~ - Non-Conv                       & 28.86 & 13.25 & 18.00 &  0.58 \\
~ - Conv                       & 27.25 & 12.65 & 18.31 &  0.58 \\
~ - High-Conv                       & 28.83 & 13.25 & 17.71 &  0.72 \\
\addlinespace[0.05cm]
GPT-4o C-R            & 27.54 & 11.44 & 16.06 &  1.73 \\
GPT-4o U-R              & 26.33 & 10.23 & 15.75 & -0.30 \\
\addlinespace[0.05cm]
Llama-3-8B-INT                 & 30.62 & 12.71 & 19.28 &  0.11 \\
Llama-8B + GPT4o C-R    & 29.83 & 12.18 & 18.00 &  0.14 \\
Llama-8B + GPT4o U-R    & 28.42 & 11.96 & 18.86 &  0.30 \\
Llama-8B + OSD-R         & 27.70 & 12.71 & 19.91 &  0.73 \\

%\hline
\end{tabular}
}
\caption{Summary of surprisal statistics across all cases of word reuse. Lower surprisal score means better coherence w.r.t. usage contexts.}
\label{tab:reuse_source_summary}
\end{table}

\paragraph{Novelty.} Table~\ref{tab:novelty_stats} shows the mean novelty scores computed over sets of slang usages. Both GPT-4o and Llama achieve high mean novelty scores across all settings, suggesting that LLMs can consistently generate more semantically divergent slang usages. Notably, human-generated OSD usages exhibits the lowest mean novelty but significantly higher dispersion indicated by high standard deviation and IQR. Our results suggest that human speakers generate slang usages in a much wider creative spectrum with a relatively loosely defined level of creativity attached to the use of slang. Meanwhile, slang reuse generated by the LLMs tend to cluster around a specific level of creativity that's more creative than the average human-attested usage.

\paragraph{Surprisal.} We measure the surprisal scores using Gemma-2-9b-Instruct~\cite{gemma2} as a judge~\cite{zheng23}, which is not a member of neither the GPT or Llama model family to ensure objectivity.
Table~\ref{tab:reuse_source_summary} shows the surprisal values. We find the surprisal values to be high across the board, indicating that machine generated slang shows nuanced control over contextual surprisal similar to those found in human usages. We find that GPT-4o in the controlled setting produced slang usages that are slightly more coherent to the context compared to humans, while the uncontrolled Llama model generates less contextually coherent usages despite having a similar level of novelty.

Overall, our results suggest that while LLMs are capable of generating creative slang usages similar to how humans do, the larger GPT-4o model tends to prefer generations that are more creative in certain aspects (i.e., morphological complexity, semantic novelty) than others (i.e., morphological coherence, contextual surprisal). In Appendix~\ref{Additional Examples}, we show example slang usages generated by GPT-4o with high and low creativity under each metric we considered in this section.
%Overall, the results suggest that while LLMs are able to produce reinterpretations that are both novel and contextually plausible, human-generated slang displays greater variability, reflecting a broader creative spectrum not fully captured by LLMs. These findings highlight both the current capabilities and limitations of LLMs on generating creative language.

\subsection{Informativeness} \label{sec_informativeness}

% \subsubsection{Informativeness via distillation}

In this section, we ask the question of whether machine-generated slang shares a similar level of informativeness when used as examples. If an LLM truly captures the nuanced generative structure behind slang usages, we would expect the machine-generated slang usages to be as informative as human generated samples.
To do this, we compare the informativeness of GPT-4o generated samples with human-generated slang under a distillation experiment~\cite{hinton15}. Specifically, we use Llama-3-8B-Instruct as the student model and finetune it using slang usages from each source. 

% While probing the capabilities of GPT-4o in handling slang, we pursue a central research question: Can knowledge of non-literal language use—particularly in terms of coinage creativity, reuse creativity, and lexical knowledge—be transferred from a teacher model to a student model? To explore this, we use Meta's LLaMA-8B-Instruct model~\cite{llama3modelcard} as the raw (i.e., pretrained but unfine-tuned) student model.

% Beyond measuring knowledge transfer, a second goal of this study is to examine whether the student model can learn the \textit{generation patterns} exhibited by GPT-4o—particularly those that implicitly reflect human slang knowledge. We are especially interested in identifying which features of GPT-4o's outputs are most helpful for enabling a smaller model to acquire human-like lexical creativity and non-literal understanding.

To ensure valid comparison and evaluation, we construct a balanced sample of slang usages with 1,000 training examples in each partition. We begin the sampling procedure by first splitting the entire OSD dataset into an 80-20 split for training and testing respectively. From the OSD training set, we randomly sample 1,000 examples for training. We also ensure that the 1,000 examples correspond to different sense clusters to maintain semantic diversity. For the GPT-4o generated partitions with controlled generation, we sample usages generated from examples in the OSD training set to avoid data contamination.
We obtain three samples from OSD with 1,000 examples in each, corresponding to free-form (OSD-F), coinage (OSD-C) and reuse (OSD-R) respectively. We also obtain 1,000 examples for each of the machine-generated partitions outlined in Table~\ref{tab:generation_schema}.
An additional 80-20 split is then applied to each 1{,}000-example set to create the final training and validation partitions used for model fine-tuning using LoRA~\cite{hu22}. Detailed experimental setup and prompts can be found in Appendix~\ref{app-compft} and \ref{sec:finetunecorpusSTrucutre}.

\subsubsection{Informing creativity}

We first finetune the Llama-8B model on either cases of coinage or reuse from both OSD and GPT-4o generated usages to examine the effect of fine-tuning on the model's creativity. Using the finetuned models, we perform uncontrolled generation of 1,000 samples from each model and repeat the experiments in Section~\ref{sec-creativity}. Table~\ref{tab:source_coinage_mean_std} and \ref{tab:coherence_scores} show the results for coinages and Table~\ref{tab:novelty_stats} and \ref{tab:reuse_source_summary} show the results for reuses.

we observe a mild signal for the transfer of coinage creativity: When Llama-8B is fine-tuned on coinages from either GPT-4o or OSD, we observe a significant increase in morphological complexity, more so when finetuned on GPT-generated slang. It shows that learning on more morphologically complex examples generated by GPT-4o does steer the student model toward generating more morphologically complex slang terms. Meanwhile, fine-tuning the model on OSD-generated entries makes the generations less coherent while fine-tuning on GPT-4o's uncontrolled generations makes them more coherent, once again showing that the fine-tuned model is being steered toward mimicking the preference of the teacher model.

When fine-tuned on reuse cases, we observe that while the Llama-generated slang attains a comparable increase in semantic novelty, the level of surprisal decreases across all cases, particularly when the model is fine-tuned on OSD slang. Interestingly, although slang usages from GPT-4o were more coherent w.r.t. the usage contexts, such knowledge does not transfer well in the fine-tuning process.

Overall, our results show that both human and machine-generated slang can provide informative information in some dimensions of creativity by steering the smaller model toward mimicking the larger model's preferences. It is important to note here that using large models such as GPT-4o as a teacher will propagate many of its biases into the smaller student model and caution should be exercised in scenarios where we want the models to faithfully represent human knowledge.

\begin{table*}[t]
\centering
% \resizebox{\textwidth}{!}{%
\begin{tabular}{lrrr}
% \hline
\textbf{Model} & \textbf{Task 1 (Acc)} & \textbf{Task 2 (Acc)} & \textbf{Task 3 (Sim)} \\%& \textbf{Task 4 (Acc)} \\
\hline
\addlinespace[0.1cm]
[OSD] \\
\addlinespace[0.05cm]
GPT-4o & 0.959 ± 0.003 & 0.989 ± 0.004 & 0.520 ± 0.001 \\%& 0.844 ± 0.004 \\
\addlinespace[0.05cm]
Llama-3-8B-INT              & 0.891 ± 0.001 & 0.910 ± 0.000 & 0.471 ± 0.001 \\%& 0.594 ± 0.000 \\
Llama-8B + GPT-4o C-F   & 0.886 ± 0.000 & 0.913 ± 0.001 & 0.494 ± 0.000 \\%& 0.590 ± 0.000 \\
Llama-8B + GPT-4o U-F   & 0.889 ± 0.001 & 0.914 ± 0.000 & 0.486 ± 0.000 \\%& 0.576 ± 0.000 \\
Llama-8B + OSD-F        & 0.882 ± 0.000 & 0.914 ± 0.000 & 0.500 ± 0.000 \\%& 0.604 ± 0.000 \\
\addlinespace[0.05cm]
\hline
\addlinespace[0.1cm]
[OpenSubtitles-Slang] \\
\addlinespace[0.05cm]
GPT-4o                & 0.965 ± 0.001 & 0.913 ± 0.004 & 0.501 ± 0.001 \\%& 0.796 ± 0.008 \\
\addlinespace[0.05cm]
Llama-3-8B-INT              & 0.928 ± 0.000 & 0.844 ± 0.000 & 0.464 ± 0.000 \\%& 0.603 ± 0.001 \\
Llama-8B + GPT-4o C-F   & 0.924 ± 0.000 & 0.838 ± 0.000 & 0.481 ± 0.000 \\%& 0.581 ± 0.001 \\
Llama-8B + GPT-4o U-F   & 0.922 ± 0.000 & 0.852 ± 0.000 & 0.475 ± 0.000 \\%& 0.582 ± 0.002 \\
Llama-8B + OSD-F        & 0.926 ± 0.000 & 0.828 ± 0.000 & 0.487 ± 0.001 \\%& 0.574 ± 0.000 \\

% \hline
\end{tabular}
% }
\caption{Performance of models across three evaluation tasks over 10 runs. Task 1 (Generation) and 2 (Interpretation) are measured by accuracy and Task 3 (Free-form interpretation) is measured by semantic similarity using SBERT.}
\label{tab:task_scores_wide}
\end{table*}

\subsubsection{Informing knowledge}

We now examine the informativeness of slang usages for downstream tasks that require structural understanding of slang. We evaluate the performance of the vanilla and fine-tuned Llama models on both slang generation and slang interpretation:
% While the previous section focuses on the creative potential of LLMs in coining or reinterpreting slang, this section 
% examines the extent to which models possess systematic knowledge of existing slang expressions. We design four diagnostic tasks to evaluate different facets of the model’s slang understanding, including lexical inference, semantic disambiguation, and usage-based reasoning.

\textbf{Task 1 - Generation} (\textit{Cloze + Definition $\rightarrow$ Word}) evaluates the model's ability to infer a slang term given a masked usage sentence (i.e., the slang term is masked out) and its corresponding definition. The model is given four choices and asked to pick the correct word choice.

\textbf{Task 2 - Interpretation} (\textit{Word + Usage $\rightarrow$ Definition}) presents a slang word in a usage sentence and asks the model to select the correct definition among four choices.

\textbf{Task 3 - Free-form interpretation} (\textit{Word + Usage $\rightarrow$ Definition (Generation)}) employs a similar setup as the interpretation task but here the model needs to generate a definition sentence for the meaning of the slang. The generated definitions are evaluated by their semantic similarity to the ground-truth definition sentence using SBERT.

% \textbf{Task 4} (\textit{Cloze $\rightarrow$ Word}) supplies only a masked usage sentence, asking the model to recover the missing slang term without access to a definition. This tests the model’s capacity for pragmatic inference and fluency in slang usage patterns.

% \begin{table}[t]
% \centering
% \resizebox{\columnwidth}{!}{%
% \begin{tabular}{lcccc}
% \hline
% \textbf{Model} & \textbf{Task 1 (Acc)} & \textbf{Task 2 (Acc)} & \textbf{Task 3 (Sim)} & \textbf{Task 4 (Acc)} \\
% \hline
% GPT-4o                     & 0.9600 & 0.9100 & 0.5051 & 0.8280 \\
% LLaMA-8B                  & 0.9220 & 0.8440 & 0.4795 & 0.5840 \\
% LLaMA-8B-INT F.t. C-F     & 0.9240 & 0.8320 & 0.4811 & 0.5580 \\
% LLaMA-8B-INT F.t. U-F     & 0.9180 & 0.8440 & 0.4770 & 0.5760 \\
% LLaMA-8B + OSD            & 0.9180 & 0.8260 & 0.4828 & 0.5600 \\
% \hline
% \end{tabular}
% }
% \caption{Performance of models across four evaluation tasks on the OpenSubtitle dataset}
% \label{tab:Opensubtitle_dataset_task_scores}
% \end{table}

% \paragraph{Overall Trends.}
%\paragraph{Results}
We evaluate both the off-the-shelf GPT-4o and Llama-3-8B-Instruct model with fine-tuned Llama models on all three tasks. We construct evaluation datasets using both OSD and OpenSub-Slang, a benchmark dataset on slang curated by \citet{sun24} using the OpenSubtitles corpus~\cite{lison16}. 
% Talk about how you set up the experiment data. How many test entries are there in each dataset.
Specifically, OSD evaluates informativeness of the generated usages in an intrinsic setting where the evaluation data shares the same data distribution as the entries used to both fine-tune Llama and control GPT-4o's generation. OpenSub-Slang, on the other hand, provides an extrinsic evaluation.
See Appendix~\ref{sec:Slang Exam Question Generation} for detailed experiment setup. 

Table~\ref{tab:task_scores_wide} summarizes the results.
GPT-4o consistently outperforms Llama models across all tasks, reflecting its robust knowledge on slang. The vanilla Llama model lags behind GPT-4o by a significant margin. We observe that finetuning on both OSD and GPT-4o generated slang yields no or minimal performance gain on tasks 1 and 2, suggesting that the student models are not able to effectively leverage the knowledge to make predictions. On Task 3, finetuning on both data sources improves the quality of the generated definitions, arguably because the model is better guided toward writing definition sentences that better conform with the dictionary style. In this case, we find human-generated OSD to be more informative compared to slang entries generated by GPT-4o. Overall, while smaller Llama models can sometimes benefit from the transfer of knowledge, the degree of improvement is task-sensitive and often constrained.

\section{Conclusion}

We have presented the first systematic comparison between human and machine-generated slang usages. Our results suggest that while LLMs such as GPT-4o achieve strong results on a wide range of evaluative tasks involving slang, structural knowledge about slang encoded in these models show notable distinctions when compared to real usages from humans. 
%Specifically, we show that while models can generate plausible slang usages, these models 
Specifically, LLMs show strong preferences toward certain characteristics and creative qualities, and such preferences can affect how the generated usages inform their users. Our findings suggest that although LLMs are capable of processing slang in ways that reflect many aspects of human knowledge, they have not yet fully captured nuanced structures in human slang usage.
% are biased toward certain modes of generation. At the same time, machine-generated slang tend to be more creative but span a less diverse spectrum of creativity compared to slang generated by humans. Finally, we discover that machine-generated slang are less informative compared to human-generated slang evident by worse downstream performance when used to distill smaller models.
%Our findings suggest that LLMs' perceptions of slang are notably different from those reflected by human-authored dictionaries, 

%are capable of performing discriminative tasks involving slang but are still lacking when used for complex tasks that rely on the informativeness of its generations, such as 

\section*{Limitations}
Due to computational budget constraints, we limit our evaluation of large language models to only Llama-8B-INT and GPT-4o. 
Although this combination captures a representative sample of both open and black-box commercial models, they do not fully capture the diverse landscape of contemporary LLMs. Ideally, we would like to expand our analysis to include a broader range of commercial systems to better understand how different models behave, as well as running large variants of open models such as Llama-70B.
%Although these models provide meaningful contrasts—one being open and fine-tuned, the other being a commercial general-purpose model—they do not capture the full diversity of contemporary LLMs. Ideally, we would expand our analysis to include a broader range of commercial systems to better understand how different models display different slang generation and interpretation. 

The scope of our study is also confined to English slang where both the human and machine-generated slang entries are only in English. Slang is inherently a cultural and multilingual phenomenon, thus evaluating how models handle slang in other languages/dialects remains an important direction for future work. Addressing these gaps would help assess whether the interpretative results we represented can be generalized across linguistic and cultural boundaries.

In this study, we focus on evaluating the models using widely adopted quantitative evaluation metrics that can be objectively and consistently applied to both human and machine-generated slang usages. We acknowledge that an evaluation grounded solely in quantitative metrics may not fully capture the richness and nuance of slang and would encourage future work to compliment our evaluations with human studies.

Lastly, we acknowledge that the use of slang dictionaries to represent human-generated slang forgo some of the more dynamic aspects of slang usages. In our work, we focus on finding potential similarities and differences between linguistically motivated aspects of slang that are well-defined for both machine and human-generated slang usages. For LLMs, we would argue that the slang usages are generated purely from learned statistical patterns and do not reflect real world experiences like those generated by humans. Although the use of slang dictionaries constrains the language in a more static view, it is consistent with what is produced by LLMs (where the dynamic aspects do not exist in the first place) and allows us to make objective comparisons in a controlled experimental setting.

\section*{Ethics Statement}
We acknowledge that many slang usages collected from dictionaries express taboo concepts. In the main text, we mask out certain example words that are deemed inappropriate and present the full results in the Appendix.
All slang usages shown in the examples were taken verbatim from the original data source and do not reflect opinions of the authors and their affiliated organizations.
Discretion is advised when viewing the examples in the Appendix and using the collected datasets.

We have been granted written permission from the author of The Online Slang Dictionary to use it for personal research purposes.

We used AI assistants to expedite the coding process. All code snippets produced by AI assistants were verified by the first author before they were incorporated. For writing, we only used AI assistants to check grammar.

\section*{Acknowledgments}

We thank the anonymous ARR reviewers and chairs for their constructive comments and suggestions. We also thank Walter Rader for permissions to use The Online Slang Dictionary for our research.

% This document has been adapted
% by Steven Bethard, Ryan Cotterell and Rui Yan
% from the instructions for earlier ACL and NAACL proceedings, including those for
% ACL 2019 by Douwe Kiela and Ivan Vuli\'{c},
% NAACL 2019 by Stephanie Lukin and Alla Roskovskaya,
% ACL 2018 by Shay Cohen, Kevin Gimpel, and Wei Lu,
% NAACL 2018 by Margaret Mitchell and Stephanie Lukin,
% Bib\TeX{} suggestions for (NA)ACL 2017/2018 from Jason Eisner,
% ACL 2017 by Dan Gildea and Min-Yen Kan,
% NAACL 2017 by Margaret Mitchell,
% ACL 2012 by Maggie Li and Michael White,
% ACL 2010 by Jing-Shin Chang and Philipp Koehn,
% ACL 2008 by Johanna D. Moore, Simone Teufel, James Allan, and Sadaoki Furui,
% ACL 2005 by Hwee Tou Ng and Kemal Oflazer,
% ACL 2002 by Eugene Charniak and Dekang Lin,
% and earlier ACL and EACL formats written by several people, including
% John Chen, Henry S. Thompson and Donald Walker.
% Additional elements were taken from the formatting instructions of the \emph{International Joint Conference on Artificial Intelligence} and the \emph{Conference on Computer Vision and Pattern Recognition}.

\bibliography{custom}
\bibliographystyle{acl_natbib}

\newpage

\appendix

\section{Techniques for improving generative diversity} \label{app-gendiver}

We evaluated the effectiveness of ~\citet{chen24} through an ablation test under default sampling parameters: temperature $= 1.0$ and top-$p = 1.0$. Comparing generation diversity in terms of the number of unique $(w, \text{sense})$ pairs across 1,000 generations under the U-F setting, we observed minimal improvement: 353 unique entries without the method versus 364 with it. This suggests that the technique does not substantially improve lexical and semantic diversity across batches.

\section{Computational setup for model distillation} \label{app-compft}

In our configuration, we set the LoRA rank to 64, the LoRA scaling factor (alpha) to 128, and applied it across all linear layers with no dropout. The fine-tuning process follows a supervised fine-tuning (SFT) setup with a batch size of 32 and a maximum gradient norm of 1.0. A cosine learning rate scheduler was employed with an initial learning rate of $1\mathrm{e}{-5}$ and no warm-up phase.

The number of training epochs was set to 5. Based on empirical observation, our models typically converges within 5 epochs. We found that extending training beyond 5 epochs yielded minimal returns in performance. Two Nvidia A6000 GPUs were used and each fine-tuning task took 16 mins. 

\section{Experiment setup for downstream tasks} \label{sec:Slang Exam Question Generation}

%To construct a reliable evaluation benchmark, we collect all data points without data leakage. The test split from OSD and the full set of entries from OpenSubtitles each serve as roles of within-domain and out-of-domain test sets specifically. 
For each task, we sample 500 evaluation examples from each source. The OSD examples are sampled from the OSD test set (described in Section~\ref{sec_informativeness}) and the OpenSub-Slang examples randomly sampled from the dataset.

\paragraph{Task 1 - Generation}  
For every slang entry, we mask the target slang word in its usage example. The original word is treated as the correct answer, and three incorrect options are sampled randomly from the remaining entries in the test set. These four options are then randomly assigned to labels A, B, C, and D, with the correct label recorded. The definition and masked usage are combined with the prompt template (shown in Section~\ref{sec:task1prompt}) to generate a multiple-choice question.

\paragraph{Task 2 - Interpretation}  
Given a slang word and its usage context, we generate a multiple-choice question asking for its correct definition. The ground truth definition serves as the correct answer, while three incorrect definitions are randomly sampled from other entries in the test set. The options are randomly ordered and labeled A–D. The question prompt follows the format shown in Section~\ref{sec:task2prompt}.

\paragraph{Task 3 - Free-form interpretation}  
For each slang word and its usage example, we generate a free-form question asking the model to write an appropriate definition. The input prompt structure is specified in Section~\ref{sec:task3prompt}, and evaluation is conducted by measuring semantic similarity (using SBERT) between the generated and ground-truth definitions.

\newpage

\section{Prompt used to generate slang usages}
\label{app:slang-prompt}
\subsection{U-F Prompt}
\begin{lstlisting}
You are a creative slang dictionary generator and here is the definition of slang:
A slang is a vocabulary (words, phrases, and linguistic usages) of an informal register, common in everyday conversation but avoided in formal writing and speech.
It also often refers to the language exclusively used by the members of particular in-groups in order to establish group identity, exclude outsiders, or both. 
Generate novel slang usages in English. 

The json structure must be:
{
  "word": [],         // An array of the slang terms
  "definition": [],   // An array of corresponding definitions
  "usage_context": [] // An array of arrays, where each array has usage examples for that slang
}

- Keep the arrays aligned so the i-th element in "word", "definition", and "usage_context" all refer to the same slang and same length.
Remember: 
- 'word' is the slang term.
- 'definition' is a short explanation.
- 'usage_context' should include 1-2 example sentences containing the slang term.

Now, generate {number_of_slang} entries.
The current dictionary already contains: [{existing_words}], do not repeat any of these.
\end{lstlisting}
\newpage

\subsection{U-R Prompt}
\begin{lstlisting}
You are a creative slang dictionary generator and here is the definition of slang:
A slang is a vocabulary (words, phrases, and linguistic usages) of an informal register, common in everyday conversation but avoided in formal writing and speech.
It also often refers to the language exclusively used by the members of particular in-groups in order to establish group identity, exclude outsiders, or both.
Generate novel slang usages in English.
'Generate' means taking existing English words and assigning them novel meanings to create novel slang, do not make up words.

The json structure must be:
{
  "word": [],         // An array of the slang terms
  "definition": [],   // An array of corresponding definitions
  "usage_context": [] // An array of arrays, where each array has usage examples for that slang
}

- Keep the arrays aligned so the i-th element in "word", "definition", and "usage_context" all refer to the same slang and same length.
Remember:
- 'word' is the slang term.
- 'definition' is a short explanation.
- 'usage_context' should include 1-2 example sentences containing the slang term.

Now, generate {number_of_slang} entries.
The current dictionary already contains: [{existing_words}], do not repeat any of these.
\end{lstlisting}
\newpage
\subsection{U-C Prompt}
\begin{lstlisting}
You are a creative slang dictionary generator and here is the definition of slang:
A slang is a vocabulary (words, phrases, and linguistic usages) of an informal register, common in everyday conversation but avoided in formal writing and speech.
It also often refers to the language exclusively used by the members of particular in-groups in order to establish group identity, exclude outsiders, or both.
Generate novel usages in English. 
'Generate' means creating novel words that do not exist in the conventional English lexicon.

The json structure must be:
{
  "word": [],         // An array of the slang terms
  "definition": [],   // An array of corresponding definitions
  "usage_context": [] // An array of arrays, where each array has usage examples for that slang
}

- Keep the arrays aligned so the i-th element in "word", "definition", and "usage_context" all refer to the same slang and same length.
Remember:
- 'word' is the slang term.
- 'definition' is a short explanation.
- 'usage_context' should include 1-2 example sentences containing the slang term.

Now, generate {number_of_slang} entries.
The current dictionary already contains: [{existing_words}], do not repeat any of these.
\end{lstlisting}

\newpage
\subsection{C-F Prompt}
\begin{lstlisting}
You are a creative slang dictionary generator and here is the definition of slang:
A slang is a vocabulary (words, phrases, and linguistic usages) of an informal register, common in everyday conversation but avoided in formal writing and speech.
It also often refers to the language exclusively used by the members of particular in-groups in order to establish group identity, exclude outsiders, or both.
Generate novel slang usages in English to express the definition: {definition}.

The json structure must be:
{
  "word": [],         // An array of the slang terms
  "definition": [],   // An array of corresponding definitions
  "usage_context": [] // An array of arrays, where each array has usage examples for that slang
}

- Keep the arrays aligned so the i-th element in "word", "definition", and "usage_context" all refer to the same slang.
Remember: 
- 'word' is the slang term.
- 'definition' is the definition that is given.
- 'usage_context' should include at least 1-2 example sentences containing the slang term.

Now, generate {number_of_slang} entries.
The current dictionary already contains: [{existing_words}], do not repeat any of these.
\end{lstlisting}
\newpage
\subsection{C-R Prompt}
\begin{lstlisting}
You are a creative slang dictionary generator and here is the definition of slang:
A slang is a vocabulary (words, phrases, and linguistic usages) of an informal register, common in everyday conversation but avoided in formal writing and speech.
It also often refers to the language exclusively used by the members of particular in-groups in order to establish group identity, exclude outsiders, or both.
Generate novel slang usages in English to express the definition: {definition}. 
'Generate' means taking existing English words and assigning them the meaning to create novel slang, do not make up words.

The json structure must be:
{
  "word": [],         // An array of the slang terms
  "definition": [],   // An array of corresponding definitions
  "usage_context": [] // An array of arrays, where each array has usage examples for that slang
}

- Keep the arrays aligned so the i-th element in "word", "definition", and "usage_context" all refer to the same slang and same length.
Remember:
- 'word' is the slang term.
- 'definition' is a short explanation.
- 'usage_context' should include 1-2 example sentences containing the slang term.

Now, generate {number_of_slang} entries.
The current dictionary already contains: [{existing_words}], do not repeat any of these.
\end{lstlisting}
\newpage
\subsection{C-C Prompt}
\begin{lstlisting}
You are a creative slang dictionary generator and here is the definition of slang:
A slang is a vocabulary (words, phrases, and linguistic usages) of an informal register, common in everyday conversation but avoided in formal writing and speech.
It also often refers to the language exclusively used by the members of particular in-groups in order to establish group identity, exclude outsiders, or both.
Generate novel slang usages in English to express the definition: {definition}.
'Generate' means creating novel words that do not exist in the conventional English lexicon.

The json structure must be:
{
  "word": [],         // An array of the slang terms
  "definition": [],   // An array of corresponding definitions
  "usage_context": [] // An array of arrays, where each array has usage examples for that slang
}

- Keep the arrays aligned so the i-th element in "word", "definition", and "usage_context" all refer to the same slang and same length.
Remember:
- 'word' is the slang term.
- 'definition' is a short explanation.
- 'usage_context' should include 1-2 example sentences containing the slang term. 

Now, generate {number_of_slang} entries.
The current dictionary already contains: [{existing_words}], do not repeat any of these.
\end{lstlisting}

\newpage
\section{Task prompts}
\subsection{Task 1 – Generation}\label{sec:task1prompt}
\begin{lstlisting}
You are given a slang usage where the slang word has been masked with a blank (___), and a definition of that slang word.

Your task is to choose the correct slang word from the four options provided.

Usage:
{masked_usage}

Definition:
{definition}

Options:
A. {A}
B. {B}
C. {C}
D. {D}

Respond with a JSON object in the following format:
{
  "answer": "your answer in a single letter chosen from the options"
}

Only output the JSON object. Do not include any explanation.
\end{lstlisting}
\newpage
\subsection{Task 2 – Interpretation}\label{sec:task2prompt}
\begin{lstlisting}
You are given a slang word and a sentence showing how it's used in context.

Your task is to choose the correct definition of the slang word from the four options below.

Word:
{word}

Usage:
{usage}

Options:
A. {A}
B. {B}
C. {C}
D. {D}

Respond with a JSON object in the following format:
{
  "answer": "your answer in a single letter chosen from the options"
}

Only output the JSON object. Do not include any explanation.
\end{lstlisting}
\newpage
\subsection{Task 3 – Free-form interpretation}\label{sec:task3prompt}
\begin{lstlisting}
You are given a slang word and a sentence showing how it is used in context.

Your task is to write a concise definition of the slang word as it is used in this context.

Word:
{word}

Usage:
{usage}

Respond with a JSON object in the following format:
{
  "answer": "your concise definition here"
}

Only output the JSON object. Do not include any explanation.
\end{lstlisting}

\subsection{Fine-tune corpus structure} \label{sec:finetunecorpusSTrucutre}
\begin{lstlisting}
Slang word: {word}\n
Defination: {definition}\n
Usage: {usage_context}\n

\end{lstlisting}
% \newpage
% \subsection{Task 4 – Cloze Only \textrightarrow Word}
% \begin{lstlisting}
% You are given a slang usage where the slang word has been masked with a blank (___).

% Your task is to choose the correct slang word from the four options provided.

% Usage:
% {masked_usage}

% Options:
% A. {A}
% B. {B}
% C. {C}
% D. {D}

% Respond with a JSON object in the following format:
% {
%   "answer": "your answer in a single letter chosen from the options"
% }

% Only output the JSON object. Do not include any explanation.
% \end{lstlisting}

\newpage
\section{Slang generation psudocode}\label{sec:slangGenerationPsudocode}
\subsection{Uncontrolled generation}
\begin{algorithm}[h]
\label{alg:slang_generation_uncon}
\KwIn{\\
    Target number of slang entries $N$;\\
    Existing slang entry set $\mathcal{E}$, where each entry is a unique tuple $(w, \text{senses})$;\\
    Generation mode $m \in \{\texttt{Freeform}, \texttt{reuse}, \texttt{coinage}\}$;
}
\KwOut{\\An expanded slang set $\mathcal{E}$ with $|\mathcal{E}| = N$ unique entries}

\While{$|\mathcal{E}| < N$}{
    Construct a prompt based on mode $m$ and query the language model to generate candidate entries $\mathcal{C}$\;
    \ForEach{$(w, \text{sense}, \text{cxt}) \in \mathcal{C}$}{
        Classify $w$ as either \texttt{coinage} or \texttt{reuse} using Wiktionary\;
        \If{$m \neq \texttt{Freeform}$ \textbf{and} the classification does not match $m$}{
            \textbf{continue}
        }
        \If{$(w, \text{sense}) \notin \mathcal{E}$}{
            Add $(w, \text{sense}, \text{cxt})$ to $\mathcal{E}$\;
        }
    }
}
\Return{$\mathcal{E}$}
\vspace{0.25cm}
\caption{Algorithm for uncontrolled slang generation}
\end{algorithm}

\newpage
\subsection{Controlled generation}
\begin{algorithm}[h]

\label{alg:slang_generation_con}
\KwIn{\\
    Existing slang dictionary $D = \{(w, \text{sense}, \text{cxt})\}$ grouped by word sense;\\
    Generation mode $m \in \{\texttt{Freeform}, \texttt{reuse}, \texttt{coinage}\}$;\\
    Initial slang set $\mathcal{E}$ containing previously generated entries.
}
\KwOut{
    A language model-generated dictionary $D'$ with $|D'| = |D|$
}

Initialize $D' \leftarrow \emptyset$\;

\ForEach{group $d \subseteq D$ corresponding to a unique word sense}{
    Initialize local slang set $\mathcal{E}_{\text{group}} \leftarrow \emptyset$\;

    \While{$|\mathcal{E}_{\text{group}}| < |d|$}{
        Construct a prompt using word sense metadata of $d$ and generation mode $m$\;
        Query the language model to generate candidate entries $\mathcal{C}$\;

        \ForEach{$(w, \text{sense}, \text{cxt}) \in \mathcal{C}$}{
            Classify $w$ as either \texttt{coinage} or \texttt{reuse} using Wiktionary\;

            \If{$m \neq \texttt{Freeform}$ \textbf{and} classification $\neq m$}{
                \textbf{continue} \tcp*{Reject mismatched mode}
            }

            \If{$(w, \text{sense}) \notin \mathcal{E}_{\text{group}} \cup d$}{
                Add $(w, \text{sense}, \text{cxt})$ to $\mathcal{E}_{\text{group}}$\;
            }
        }
    }

    Append $\mathcal{E}_{\text{group}}$ to $D'$\;
}

\Return{$D'$}
\vspace{0.25cm}
\caption{Algorithm for controlled slang generation}
\end{algorithm}

\newpage
\section{Coinage category classification psudocode}\label{sec:CoinageCategoryClassificationPsudocode}
\begin{algorithm}[h]
\label{alg:coinage_classification}
\KwIn{\\
    A dataset of slang words $\mathcal{D}$, where each word $w$ is a candidate coinage;\\
    Wiktionary index $\mathcal{W}$ mapping known words to definitions;\\
    A trained Morfessor segmentation model $\mathcal{M}$;
}
\KwOut{\\A labeled dataframe $\mathcal{T}$ where each word is assigned a category label in \texttt{\{Compound, Blend, Other\}}}

Initialize empty record list $\mathcal{T}$\;

\ForEach{source group $(s, W_s) \in \mathcal{D}$}{
    \ForEach{word $w \in W_s$}{
        Segment $w$ into subword units $S = \mathcal{M}.\texttt{segment}(w)$\;
        \uIf{$|S| \geq 2$ }{
        \uIf{all $s_i \in S$ are exact matches in $\mathcal{W}$}{
            Label $w$ as \texttt{Compound}
        }
        \uElseIf{$s_1 \in S$ is a preffix of some $w' \in \mathcal{W}$ and $s_{-1} \in S$ is a suffix of some $w' \in \mathcal{W}$ }{
            Label $w$ as \texttt{Blend}
        }
        }
        \Else{
            Label $w$ as \texttt{Other}
        }
        Append $(s, w, |S|, \text{label})$ to $\mathcal{T}$
    
    }
}
\Return{$\mathcal{T}$}
\vspace{0.25cm}
\caption{Algorithm for classifying coinage types using Morfessor and Wiktionary}
\end{algorithm}

\newpage
\section{Sample data} \label{app-sampledata}

Table~\ref{tab:combined_slang} shows examples from OSD and machine-generated slang usages.

\section{Topic analysis Full table}\label{sec:Topic analysis Full table}

Table~\ref{tab:lda_all_topics} shows the full results for the topic modeling experiment described in Section~\ref{sec_characteristics}.

\section{Additional Examples}\label{Additional Examples}

Table~\ref{tab:blend_compound_other} shows examples of how different word formation strategies are employed. Table~\ref{tab:blend_compound_other} through \ref{tab:surprisal_examples} show machine-generated slang usages with high and low scores in each measure of creativity we considered in the paper.

\begin{table*}[t]
\centering
\resizebox{\textwidth}{!}{%
\begin{tabular}{@{}p{1.5cm}p{2.5cm}p{5cm}p{14cm}@{}}
% \toprule
\textbf{Source} & \textbf{Slang term} & \textbf{Definition} & \textbf{Usage context} \\
\hline
% \midrule
\addlinespace[0.1cm]
OSD & bruddah & alternate spelling of brother. & Safe, my bruddah. \\
OSD & cat off & Doing something out of the ordinary or stupid. & You cattin' off coming at me like that. Jerry went up to the girl to ask for a dance and she catted him off. \\
OSD & crop dust & to flatulate while walking through an area or by group of people. & Whoa! Smells like somebody has been crop dusting. He came in and crop dusted us. \\
OSD & cuckoo & crazy. & He's cuckoo. \\
OSD & cunt-fuck & to have vaginal sex. & My girlfriend and I got so wasted last night she asked me to cunt-fuck her. \\
\addlinespace[0.05cm]
\rowcolor{gray!20} % shaded row
C-C & zucchini zip & humorous word for a penis & He charmed everyone with tales of his zucchini zip. Expect his zucchini zip stories to get a chuckle from the crowd. \\
\rowcolor{gray!20} % shaded row
C-C & BeatBox808 & A device that produces the signature sounds of the Roland 808. & The BeatBox808 was laying down a perfect bassline for the session. His set was on fire once he incorporated the BeatBox808 into the rhythm. \\
\rowcolor{gray!20} % shaded row
C-C & AUX & Acronym for “as you understand”. & Can you give me a quick recap AUX? AUX, we're going for dinner at the usual spot. \\
\rowcolor{gray!20} % shaded row
C-C & AFUNU & Acronym for “as far as you know”. & We're still meeting up later, AFUNU. AFUNU, they haven't decided on a location yet. \\
\rowcolor{gray!20} % shaded row
C-C & baebs & An endearing abbreviated form of “babe”. & Hey baebs, want to grab some dinner tonight? How was your day, baebs? \\
\addlinespace[0.05cm]
C-F & scoot & A casual and informal way to indicate you are leaving. & Finished my work, I'm gonna scoot! It's getting late, time for me to scoot. \\
C-F & sloshed & Extremely drunk, to the point of losing control. & He was so sloshed he couldn’t even walk straight. She was sloshed after the party and had to take a cab home. \\
C-F & you bunch & A casual way of referring to everyone present or being spoken to. & You bunch better be ready for the game tonight! Where's the energy, you bunch? Let's get hyped up! \\
C-F & nugget & A small, cute term for a breast. & Her shirt was tight, revealing the outline of a nugget. A gentle pat on her nugget was met with a playful smack. \\
C-F & SAGZ & An acronym for Sex, Age, Gender, Zodiacs. & Instead of the usual small talk, she popped a SAGZ. When someone asks me for my SAGZ, it makes the chat more engaging. Try it next time! \\
\addlinespace[0.05cm]
\rowcolor{gray!20} % shaded row
C-R & day one & A best friend who has been there from the beginning. & He's my day one, always there since the beginning. I can trust her with anything; she's been my day one. \\
\rowcolor{gray!20} % shaded row
C-R & hubcap & A term for your significant other who keeps your world running smoothly. & Whenever things get hectic, I know my hubcap is there to keep everything together. She’s not just my partner, she’s my hubcap making the everyday run smoothly. \\
\rowcolor{gray!20} % shaded row
C-R & spin the bottle & To perform fellatio. & I think she's going to spin the bottle with him later. Let's see who's brave enough to spin the bottle tonight. \\
\rowcolor{gray!20} % shaded row
C-R & dig deep & To thoroughly review or research prior studies. & The student had to dig deep into past studies to find the missing link in his research. To fully understand the context, I needed to dig deep into prior academic journals. \\
\rowcolor{gray!20} % shaded row
C-R & memory jogger & An action to stir recollections or awareness. & Her old friend's visit acted like a memory jogger, bringing back countless memories. After procrastinating all day, the looming deadline was a real memory jogger. \\
\addlinespace[0.05cm]
U-C & blizzleplunk & A sudden change in direction during a walk or drive. & When he turned that corner, it was a blizzleplunk all the way! Our walk turned into a blizzleplunk after he got us lost. \\
U-C & splogboop & An unexpected delightful surprise & I found a \$20 bill on the street today, total splogboop! \\
U-C & blizzlefrost & A cold, frosty chill of excitement & The first snowflake of the season gave me such a blizzlefrost. \\
U-C & trungleflap & To haphazardly bounce or tumble & Watch out, don't trungleflap over the rug! \\
U-C & zorkmingle & A quirky social gathering & We went to a zorkmingle at Jane's place last night. \\
\addlinespace[0.05cm]
\rowcolor{gray!20} % shaded row
U-F & fluffle & A cozy group of cute or fluffy things gathered together. & The fluffle of kittens was too cute to handle. Nothing beats a fluffle of bunnies in the morning to lift your spirits. \\
\rowcolor{gray!20} % shaded row
U-F & doomscroll & The act of endlessly scrolling through bad news. & I lost two hours to doomscroll on Twitter last night. To break the cycle, I've installed an app to curb my doomscroll habit. \\
\rowcolor{gray!20} % shaded row
U-F & jugglework & The complex act of balancing multiple tasks at work. & Working in marketing means constant jugglework, especially during campaigns. She manages her jugglework skillfully, balancing three roles seamlessly. \\
\rowcolor{gray!20} % shaded row
U-F & techtime & Quality screen time for relaxation or productivity. & We've scheduled techtime to catch up on some documentaries this weekend. Let’s have a techtime session to binge some classics. \\
\rowcolor{gray!20} % shaded row
U-F & cringeflash & The rush of secondhand embarrassment from awkward memories. & I experienced a cringeflash when I remembered my failed public speech in high school. Every time that song plays, I get a cringeflash of my awkward dance moves. \\
\addlinespace[0.05cm]
U-R & backwash & The residual effects of an event or situation. & After the festival, there was a backwash of positive energy and camaraderie. The media backwash from the announcement was overwhelming. \\
U-R & cinderblock & Solid and unmovable, like firm determination. & Her determination was like a cinderblock, unyielding and strong. We need a cinderblock of confidence to get through this challenge. \\
U-R & switchblade & A quick, witty comeback or response. & After a bit of banter, Mike unleashed his switchblade that left everyone speechless. Her quick switchblade during the discussion won her a lot of applause. \\
U-R & lanternfish & Someone who is a night owl. & Kyle, the lanternfish of the group, is always busy when the rest of us are sleeping. The neighborhood knows Sam as a lanternfish because his lights are always on at midnight. \\
U-R & bathrobe & The state of feeling relaxed and at ease. & After a long week, the weekend felt like slipping into a bathrobe, comfortable and warm. Whenever I'm stressed, talking to Jenny is like putting on a bathrobe. \\
%\bottomrule
\end{tabular}%
}
\caption{Randomly sampled examples from both GPT-generated slang usages and OSD.}
\label{tab:combined_slang}
\end{table*}

\clearpage

\begin{table*}[h]
\centering
\small
\begin{tabular}{cp{0.28\linewidth}p{0.28\linewidth}p{0.28\linewidth}}

\textbf{Topic} & \textbf{OSD topic words} & \textbf{GPT-4o topic words} & \textbf{LLaMA-8B topic words} \\
%\textbf{Topic} & \textbf{OSD} & \textbf{GPT-4o} & \textbf{LLaMA-8B-INT} \\
\hline
\addlinespace[0.1cm]
1 & adjective, get, person, term, shit, acronym, think, guy, bad, dude
  & quick, unexpected, reaction, laughter, cause, change, catch, news, mind, situation
  & new, party, find, flumplen, add, decorate, group, night, idea, search \\
\rowcolor{gray!20} % shaded row
2 & go, time, fuck, want, ass, way, good, work, think, say
  & idea, give, make, plan, subtle, creativity, project, look, new, thought
  & new, need, ing, kid, catch, friend, jargle, playful, stop, love \\

3 & female, give, male, need, boy, ride, movie, face, girlfriend, little
  & excitement, sudden, unexpected, moment, energy, meeting, cause, intense, feel, room
  & feel, get, energy, feeling, dance, concert, start, crowd, party, excitement \\
\rowcolor{gray!20} % shaded row
4 & look, night, uncountable, person, go, sex, guy, cool, number, play
  & day, gentle, feel, party, light, movement, room, energy, lively, add
  & try, way, friend, good, hour, get, snurfle, work, flumplen, end \\

5 & person, man, find, save\_deletion\_legitimate\_citation, definition\_questionable, pende\_deletion, let, woman, get, come
  & surprise, leave, excitement, dance, sudden, movement, action, event, energy, unexpected
  & flumplen, look, try, situation, team, snurfle, person, new, take, room \\

\end{tabular}

\caption{Top-10 LDA topic words from both OSD, GPT-4o, and Llama-3-8B sense definitions.}
\label{tab:lda_all_topics}
\end{table*}

% \clearpage
% \newpage

\begin{table*}[h]
\centering
\begin{tabular}{p{2cm}p{3cm}p{4.5cm}p{5.5cm}}

\textbf{Category} & \textbf{Slang term} & \textbf{Slang definition} & \textbf{Usage example} \\
\hline
\addlinespace[0.1cm]
Blend
& \textbf{flombast} & To loudly express excitement
& He couldn't help but flombast when he found out the good news. \\
Blend
& \textbf{swoopaloo} & A spontaneous dance move that resembles a quick swoop
& Her sudden swoopaloo made everyone laugh. \\
\addlinespace[0.1cm]
\rowcolor{gray!20} % shaded row
Compound
& \textbf{zingfoot} & The feeling of sudden joy in the legs
& The music made her feel zingfoot, and she danced happily. \\
\rowcolor{gray!20} % shaded row
Compound
& \textbf{jamwhisk} & Chaotically combining jamming and whisking tasks
& His approach was pure jamwhisk—chaos but fun! \\
\addlinespace[0.1cm]
Other
& \textbf{snurfle} & To search for something in a distracted manner
& Can you snurfle that address book for my birthday? I keep misplacing it. \newline
I tried snurfle my keys for an hour before realizing I left them at work. \\
Other
& \textbf{wumwum}& A nonsensical phrase used to avoid commitment
& Wumwum, dude, I don't know if I can make it to the party tonight. \newline
My friend never commits to anything, always saying `wumwum' to get out of plans. \\
Other
& \textbf{EDAI} & Acronym for ``as far as I understand''
& The professor's explanation was too complex, so I just nodded along, OIAFU I had no idea what he was talking about. \newline
I love this new restaurant, EDAI the chef is amazing. \\

\end{tabular}
\caption{Examples of word formation strategies in slang coinages from GPT-4o U-C.}
\label{tab:blend_compound_other}
\end{table*}

\begin{table*}[h]
\centering

\begin{tabular}{p{2cm} p{3cm}p{3.7cm} p{6.3cm}}

\textbf{Coherence} & \textbf{Slang term \phantom{111} (Coherence score)} & \textbf{Slang definition} & \textbf{Constituent words and their definitions} \\
\hline
\addlinespace[0.1cm]
High
& \textbf{twirlpluck} (1.432)& The elegant movement of swirling and twisting.
& \textit{twirl}: to spin or rotate. \newline
  \textit{pluck}: to pull or pick something, often delicately. \\
High
& \textbf{whirlbloop} (1.422)& A rapid movement accompanied by a whooshing sound.
& \textit{whirl}: a fast spinning motion. \newline
  \textit{bloop}: an onomatopoeic sound, often for a soft pop or blip. \\
\addlinespace[0.1cm]
\rowcolor{gray!20} % shaded row
Low
& \textbf{flobberglitch} (1.158)& The process of overcoming minor technical problems with makeshift solutions.
& \textit{glitch}: a sudden, usually temporary malfunction or fault of equipment. \newline
  \textit{flobber}: to sag and wobble. \\
  \rowcolor{gray!20} % shaded row
Low
& \textbf{gorkflump} (1.154)& The state of feeling confused and delighted at the same time.
& \textit{gork}: informally, dazed or semi-conscious state. \newline
  \textit{flump}: a soft, heavy fall or collapse; a sudden plop or drop. \\

\end{tabular}
\caption{Examples of slang coinages with high and low morphological coherence scores from GPT-4o U-C.}
\label{tab:coherence_examples}
\end{table*}

\begin{table*}[h]
\centering
\begin{tabular}{p{2cm}p{3.5cm}p{4cm}p{5.5cm}}

\textbf{Complexity} & \textbf{Slang term \phantom{1111111} (\# of segments)} & \textbf{Slang Definition} & \textbf{Usage Example} \\
\hline
\addlinespace[0.1cm]
High
& \textbf{sploodlekabob} (5) & A delightful combination of foods .
& Their potluck dish was a sploodlekabob everyone wanted. \\
High
& \textbf{flimpflarble} (4) & Elegant, yet slightly comical coordination.
& Their routine included a daily flimpflarble of antics. \\
\addlinespace[0.1cm]
\rowcolor{gray!20} % shaded row
Low
& \textbf{glizzle} (1) & To glaze something with a light-hearted or humorous tone.
& After reading the joke book, he decided to glizzle his day with laughter. \\
\rowcolor{gray!20} % shaded row
Low
& \textbf{quarkflunk} (2) & An inexplicable mood that lasts the entire day.
& She woke up feeling a quarkflunk she couldn't shake off. \\

\end{tabular}
\caption{Examples of slang coinages with high and low morphological complexity scores from GPT-4o U-C.}
\label{tab:complexity_examples}
\end{table*}

\begin{table*}[h]
\centering
\begin{tabular}{p{2cm}p{3.7cm}p{4.5cm}p{4.8cm}}

\textbf{Novelty} & \textbf{Slang term \phantom{1111111} (Novelty score)} & \textbf{Slang definition} & \textbf{Usage example} \\
\hline
\addlinespace[0.1cm]
High
& \textbf{flashdrive} (1.4338) & When someone abruptly changes topics during a conversation.
& She totally flashdrived the conversation when we were talking about food and she brought up her vacation plans. \\
High
& \textbf{meatloaf} (1.4064) & A situation or project that is substantial but predictable and uninspired.
& The project was a real meatloaf, no surprises, but it got the job done. \\
\addlinespace[0.1cm]
\rowcolor{gray!20} % shaded row
Low
& \textbf{scatterbrain} (0.5801) & A person who is often forgetful or disorganized.
& Emma forgot her keys again, classic scatterbrain move. \\
\rowcolor{gray!20} % shaded row
Low
& \textbf{couch potato} (0.6378) & A person who spends a lot of time sitting and watching TV.
& He's such a couch potato, always glued to the TV. \\

\end{tabular}
\caption{Examples of slang reuses with high and low novelty scores from GPT-4o U-R.}
\label{tab:novelty_examples}
\end{table*}

\begin{table*}[h]
\centering
\begin{tabular}{p{2cm}p{3.7cm}p{3.8cm}p{5.5cm}}

\textbf{Surprisal} & \textbf{Slang term \phantom{1111111} (Surprisal score)} & \textbf{Slang definition} & \textbf{Usage example} \\
\hline
\addlinespace[0.1cm]
High
& \textbf{stepladder} (60.5) & An unexpected boost or assistance.
& As soon as we faced a challenge, our mentor appeared like a stepladder. \\
High
& \textbf{rockslide} (57.375) & An emotional or social overwhelm.
& After the conference, my email was hit with a rockslide of follow-ups. \\
\addlinespace[0.1cm]
\rowcolor{gray!20} % shaded row
Low
& \textbf{lifeline} (6.781) & A vital support system.
& Her words were a lifeline during those stormy days. \\
\rowcolor{gray!20} % shaded row
Low
& \textbf{umbrella} (6.875) & Protection or safeguard, shielding one from metaphorical rain.
& She was my umbrella during the stormy days of that job. \\

\end{tabular}
\caption{Examples of slang reuses with high and low surprisal scores from GPT-4o U-R.}
\label{tab:surprisal_examples}
\end{table*}

\end{document}